\documentclass{article}
\usepackage{ijcai19}

\usepackage{times}
\usepackage{soul}
\usepackage{url}
\usepackage[hidelinks]{hyperref}
\usepackage[utf8]{inputenc}
\usepackage[small]{caption}
\usepackage{graphicx}
\usepackage{amsmath}
\usepackage{booktabs}
\usepackage{algorithm}
\usepackage{algorithmic}
\title{Automatically Generating Macro Research Reports from a Piece of News}

\author{
Wenxin Hu$^1$\footnote{Contact Author}\and
Xiaofeng Zhang$^2$\and
Gang Yang$^3$\\
\affiliations
$^{1}$Harbin Institute of Technology, Shenzhen\\
\emails
huwenxin@stu.hit.edu.com,
zhangxiaofeng@hit.edu.com,
yanggang@stu.hit.edu.com,
}

\begin{document}

\maketitle

\begin{abstract}

Automatically generating macro research reports from economic news is an important yet challenging task. %As is known,
As we all know, it requires the macro analysts to write such reports within a short period of time after the important economic news are released. This motivates our work, i.e., using AI techniques to save manual cost. The goal of the proposed system is to generate macro research reports as the draft for macro analysts. Essentially, the core challenge is the long text generation issue. To address this issue, we propose a novel deep learning technique based approach which includes two components, i.e., outline generation and macro research report generation. %For the encoder phase, we encode the input news into features via word embedding technique. For our first step is to generate a outline of report, our second step is to mask words of outline and generate a final report. ｛这部分是直接删除？？｝
For the model performance evaluation, we first crawl a large news-to-report dataset and then evaluate our approach on this dataset, and the generated reports are given for the subjective evaluation.   

\end{abstract}

\section{Introduction \& Related Works}

Nature language generation (NLG) is a fundamental yet challenging task in the domain of Natural Language Processing (NLP). NLG has long been studied in many applications such as dialogue system \cite{Akasaki2017Chat,serban2016multiresolution}, machine translation \cite{Chiang2010Learning} and summarization \cite{see2017get,chen2018fast}. One emerging application of NLG is the macro research report generation which usually takes short news as the input and outputs long reports. 

Generally, the macro research reports could be classified into different categories such as company level research report, industry level research report and macro research report. Different type of reports may have different requirements. For macro research report, it usually requires the macro analysts or economists to release their reports as soon as possible after some important macro-economic news are released. This is very competitive and requires a high manual cost. This motivates us to propose this approach to automatically generate macro research reports from news.

Conventionally, text generation is one of the most important NLP tasks. Many research efforts have been made towards this end such as poem generation \cite{Liu2017A} and novel writing \cite{Yang2011A}. There are three main streams of generative based approaches, e.g., variational autoencoder based approach \cite{kingma2014auto-encoding}, auto-regressive based approach and generative adversarial networks based approach. Many attempts have been made to utilize VAEs \cite{bowman2016generating,miao2016neural} and GANs \cite{kusner2016gans,yu2017seqgan:} for the text generation task. However, the length of the output texts of these approaches are usually less than 40 words \cite{guo2018long}. That is, we need to seek a novel approach for this task to generate a much longer text from very short textual data. 

%After the processing and relevant analysis of massive data and daily news, operator can produce the reports. These reports often need high timelines, a lot of data collection and quick analysis. It is an important research problem that how to generate financial long research report automatically. 

%To generate a long text (up to 140 words), a simple method is to train a recurrent neural network model (RNNLM) \cite{bengio2003a} by maximizing the log-likelihood of each ground-truth word. However, the assumed prior distribution of observed terms is usually biased one and consequently results in a poor model performance. 
Recently, sequence-to-sequence model \cite{dusek2016sequence-to-sequence} is proposed and this approach is a two-phase one. In the first phase, it generates a semantic representation according to the input sentences. In the second phase, many coherent sentences are generated according to the input sentences. However, this approach has two major shortcomings. First, the length of its output is comparably short (less than 140 words) which is not suitable for the generation of long macro-research reports. Second, the intermediate output of each iteration is susceptible to the output of its previous results which results in many repeat terms in the final output. 

To cope with this challenge, we propose this novel approach to tackle the issue. If we generate long reports from very short text, there must exists many hidden information which is hard to directly model. Intuitively, macro analysts would draft outlines before writing the macro research reports. Inspired by this, we propose to learn the outlines from a pair of news and reports via deep neural networks. Then, we can generate the reports from outlines (feature dimension decreased) via a decoder process (feature dimension increased). The global loss function could then be proposed to minimize the loss from the input news to the generated reports. 

%The main contributions of our work can be summarized as follows. 
%\begin{itemize}
%    \item We carefully prepare a news-to-report dataset (in Chinese). We crawl macro news together with the macro research reports from \textit{downloadable 下载的？改成爬取的？} Web sites. This dataset could be used as the benchmark dataset for the task that generating long text from short text. 
%    \item We propose a novel deep neural network model to automatically generate long macro research reports. It is a end-to-end model which bypasses the complex modeling process and can directly output macro research reports from the input news. 
%    \item \textit{We design two stages in decoder process. Our model can produce the outline of report firstly using \textit{the summary of 删除？}the report and news. And then model can generate the final report using the outline of report and news. It can understand context of input report better.}
%\end{itemize}

\section{The Proposed Approach}

\begin{figure}[h]
\includegraphics[width=7cm, height=3.6cm]{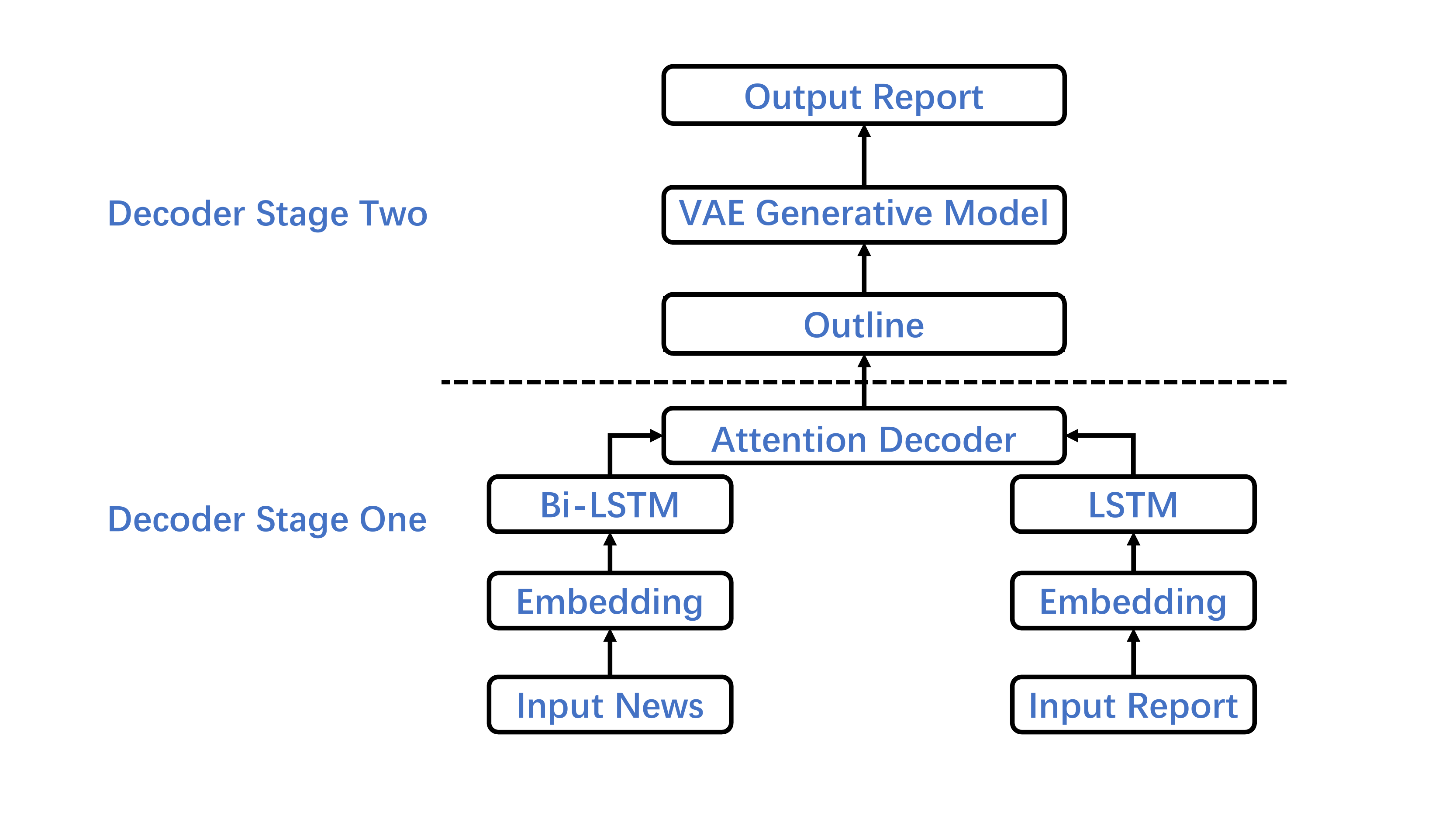}\caption{Framework of the proposed approach.}\label{fig:Framework}
\end{figure}

The framework of the proposed approach is illustrated in Figure \ref{fig:Framework}. The general idea is briefly illustrated as follows. From the bottom, we have a pair of input, e.g., the input news and the corresponding reports. Through an attention component, more important terms are kept to form the outline of this report. The input news are embedded to word vectors and are sent to a Bi-LSTM module to learn the sequence order. The learned outline is then used to generate reports. We will detail our approach in the following subsections. 

%We first briefly explain the structure of framework, which learns to generate a macro research report from a corresponding news. Then describe the process of encoder and decoder. Finally we discuss the optimization of our objective function.
%---------------------
%---------------------
%\subsection{Problem Formulation}
Let $X = \{x_1, \ldots, x_m\}$ denote the input news with $x_i$ is one of the $N$ items contained in vocabulary $V = \{v_1, \ldots, v_n\}$, $Y = \{y_1, \ldots, y_L\}$ denote the corresponding macro research reports and $O = \{o_1, \ldots, o_L\}$ denote the outline generated with $L$ as its length.

%We first use daily news X for generating the outline of report. Then we use the outline and corresponding news together for generating the final report. The outline assists news in comprehending the structure and content of report.

\subsection{Outline Generation} 
The outline is generated using the framework of sequence-to-sequence model \cite{xing2017topic}. The input news is first embedded into a word vector and then is fed into a Bi-LSTM model to acquire its hidden state $H^e = \{h_1^e, \ldots, h_t^e\}$, calculated as 
%--------------------  
\begin{equation}
    h_t^e = f_{\overrightarrow{enc}}(x_{t}, h_{t-1}^e) \| f_{\overleftarrow{enc}}(x_t, h_{t+1}^e).
\end{equation}
%--------------------
Then, the $H^e$ is used to generate the outline.

To generate the outline, we adopt LSTM network to estimate the probability distribution over the vocabulary set. It outputs the next token based on the generated summary (denoted as a low dimensional vector) of input report $Y$. At each step $t$, the current hidden state $s_t$ depends on both the token $o_{t-1}$ and the hidden state $s_{t-1}$, written as 
%--------------------
\begin{equation}
    s_t^d = f_{dec}(\hat{o}_{t-1}, s_{t-1}^d)
\end{equation}
%--------------------
To assign higher weights to more important terms, an attention mechanism is introduced here. Accordingly, the dot product of encoder hidden state $h_j^e$ and decoder hidden state $s_i^d$ is calculated as 
%-----------------------
\begin{equation}
    e_{ij} = score(h_j^e, s_i^d) = h_j^e \cdot s_i^d.
\end{equation}
%-----------------------
Then, a softmax function is employed to calculate the weights based on the scores. The results are given as %---------------------
\begin{equation}
    \alpha_{ij} = \frac{exp(e_{ij})}{\sum{exp(e_{ij})}}
\end{equation}
\begin{equation}
    c_i = \sum_1^T{\alpha_{ij} \cdot h_j^e}
\end{equation}
\begin{equation}
    \hat{s}_t = \tanh(W_c[c_t; s_t])
\end{equation}
%---------------------
where $c_i$ denote the context vector, $\hat{s}_t$ denote the attention state, and $s_t$ is the decoder hidden state. Then, the probability of the next token (term to be generated) $o_t$ could be computed as $P(o_t|o_{<t}, x) = softmax(W_c \cdot \hat{s}_t)$. The objective function of this process is given as 
%---------------------
\begin{equation}\label{eq:Obj}
    \hat{L}_{outline}(\theta_E, \theta_O; x) = \sum_{i=1}^{o}{-logP(o_i = o^*|S)}
\end{equation}
%---------------------

\subsection{Macro Research Report Generation}
\begin{figure}[h]
\includegraphics[width=7.8cm, height=3.8cm]{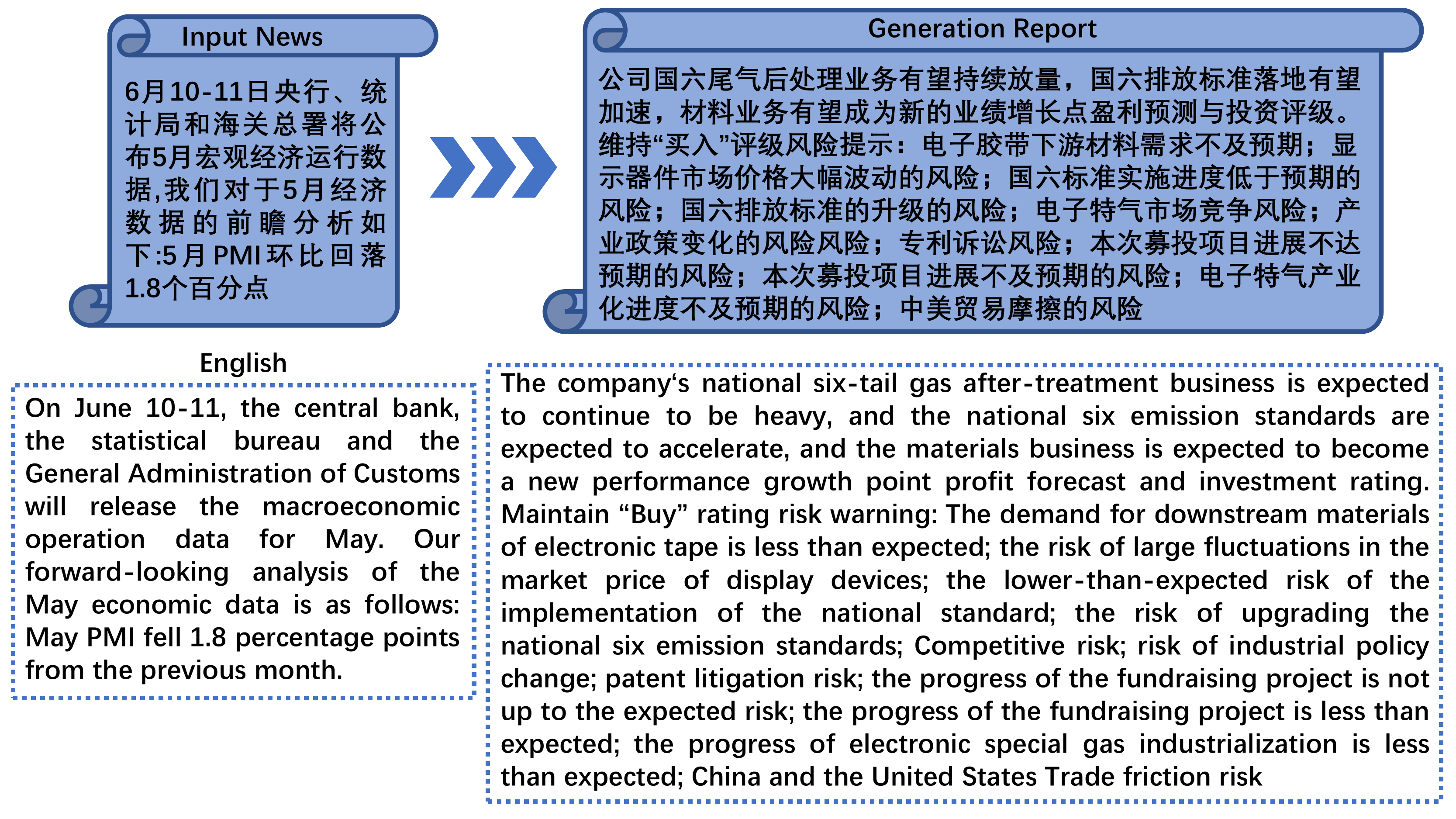}\caption{Sample results: one input news and the corresponding generated report. }\label{fig:Result}
\end{figure}

To generate the final report, we use both the news and the outline trained to generate the report. This process is also treated as a decoding process. The VAE based approach is employed as the decoding component. For the input of VAE, a mixture of news and outline is treated as the input. The original VAE model tends to learn the distribution of latent variables and samples data from this distribution. Its logarithm likelihood of sampled data $x$ is acquired by by maximizing the ELBO problem, given as $logP(X) \geq E_{q(z|x)}[logp(x|z)] - KL(q(z|x)||p(z))$.
In this stage, the decoder hidden state $h^d$ is updated as
\begin{equation}
    h_t^d = f_{dec}(\hat{y}_{t-1}, h_{t-1}^d)
\end{equation}
%--------------------
Similar objective function could be acquired, given as
%------------------------
\begin{equation}\label{eq:YObj}
    \hat{L}_{report}(\theta_R; o) = \sum_{i=1}^{y}{-logP(y_i = y^*|H)}
\end{equation}
%------------------------
%Note that both Equation \ref{eq:Obj} and \ref{eq:YObj} are local loss functions. 
At last, the global loss function of the model could be designed by considering the cost of outline generation Eq. \ref{eq:Obj} and the cost of report generation Eq. \ref{eq:YObj}. Accordingly, the global objective function is given as
%-------------------
\begin{equation}
    \hat{L}_{model}(\theta_E, \theta_O, \theta_R; x) = \hat{L}_{outline} + \hat{L}_{report}
\end{equation}
%-------------------
Experiments are evaluated on the crawled dataset, and a pair of news and the generated report are given in Figure \ref{fig:Result} for subjective evaluation. 
%---------------------
%---------------------

%sample generated reports report the evaluation criterion Cumulative BLEU which is widely adopted one in text generation task. More results are given in the demo video. The Cumulative BLEU2-5 values of our approach gain significance performance which could be considered as a rather good one. 

%The statistics of the experimental datasets are reported in Table \ref{tab:dataStat}.

%\begin{table}[!htbp]
%\renewcommand{\arraystretch}{1.2}
%    \centering
%    \caption{Cumulative BLEU scores on %different length of reports}
%    \begin{tabular}{cccccc}
%    \hline
%        len(news) & len(reports)& BLEU-2 & %BLEU-3 & BLEU-4\\
%         \hline
%         50&150& 0.183 & 0.321 & 0.425 \\
%         50&200& 0.163 & 0.298 & 0.404\\
         
%         \hline
%    \end{tabular}
%    \label{tab:dataStat}
%\end{table}

%30&150&\\
%30&200&\\

\section{Conclusion \& Future Work}
In this paper, we propose an approach which can automatically generating macro research report from economic news. This task is very challenging and we output the reports for subjective evaluations. In fact, we already computed the BLEU value which is widely adopted criterion in NLP. However, we have not implemented other state-of-the-art approaches due to the limit time. In the near future, we will implement the BLEU as well as other criteria for more approaches to evaluate model performance. 
%is just  subjective evaluation. In future work, we should enhance the inferential capability in the process of encoder-decoder. And we can try to generate longer macro research report in shorter news. Also we plan to generate different level report, such as company research report and industry research report. It is worth to rebuild the structure and generate better report.

\bibliographystyle{named}
\bibliography{ijcai19}
\bibliographystyle{ijcai19}

%\newpage{~~~}
%Requirements of the Demo Setting
%\begin{itemize}
%    \item A Poster
%    \item A Table
%    \item Internet Access
%\end{itemize}

\end{document}